\documentclass[smallextended]{svjour3} 
\smartqed  % flush right qed marks, e.g. at end of proof
\usepackage{amsmath}
\usepackage{txfonts}
\usepackage{rotating}
\usepackage{bbding}

\usepackage[style=authoryear,sorting=nyt,backend=bibtex,maxbibnames=99]{biblatex}
\AtBeginBibliography{}
\DeclareNameAlias{sortname}{last-first}
\DeclareNameAlias{default}{last-first}
\bibliography{References}
\usepackage{eso-pic}% http://ctan.org/pkg/eso-pic
\begin{document}
\AddToShipoutPictureBG*{%
  \AtPageUpperLeft{%
    \hspace{\paperwidth}%
    \raisebox{-\baselineskip}{%
      \makebox[0pt][r]{\textit{Paper accepted: Journal of Classification (Springer), expected in 2016.}
}}}}%
\title{A survey on feature weighting based K-Means algorithms}
\author{Renato Cordeiro de Amorim}
\institute{RC de Amorim \at
              Department of Computer Science, University of Hertfordshire, College Lane, Hatfield AL10 9AB, UK.\\
              Tel.: +44 01707 284345\\
              Fax: +44 01707 284115\\
              \email{r.amorim@herts.ac.uk}}
%\institute{Removed for review}

\date{Received: 21 May 2014, Revised: 18 August 2015.}
\maketitle
\begin{abstract}
In a real-world data set there is always the possibility, rather high in our opinion, that different features may have different degrees of relevance. Most machine learning algorithms deal with this fact by either selecting or deselecting features in the data preprocessing phase. However, we maintain that even among relevant features there may be different degrees of relevance, and this should be taken into account during the clustering process.

With over 50 years of history, K-Means is arguably the most popular partitional clustering algorithm there is. The first K-Means based clustering algorithm to compute feature weights was designed just over 30 years ago. Various such algorithms have been designed since but there has not been, to our knowledge, a survey integrating empirical evidence of cluster recovery ability, common flaws, and possible directions for future research. This paper elaborates on the concept of feature weighting and addresses these issues by critically analysing some of the most popular, or innovative, feature weighting mechanisms based in K-Means.

\keywords{Feature weighting \and K-Means \and partitional clustering \and feature selection.}
\end{abstract}
\section{Introduction}
%BASIC BACKGROUND AND MOTIVATION

Clustering is one of the main data-driven tools for data analysis. Given a data set $Y$ composed of entities $y_i \in Y$ for $i=1, 2, ..., N$, clustering algorithms aim to partition $Y$ into $K$ clusters $S=\{S_1, S_2, ..., S_K\}$ so that the entities $y_i \in S_k$ are homogeneous and entities between clusters are heterogeneous, according to some notion of similarity. These algorithms address a non-trivial problem whose scale sometimes goes unnoticed. For instance, a data set containing 25 entities can have approximately $4.69x10^{13}$ different partitions if $K$ is set to four \parencite{steinley2006k}. Clustering has been used to solve problems in the most diverse fields such as computer vision, text mining, bioinformatics, and data mining \parencite{vedaldi2010vlfeat,steinley2006k,jain2010data,sturn2002genesis,huang2008weighting,gasch2002exploring,mirkin2012clustering}.

Clustering algorithms follow either a partitional or hierarchical approach to the assignment of entities to clusters. The latter produces a set of clusters $S$ as well as a tree-like relationship between these clusters, which can be easily visualised with a dendogram. Hierarchical algorithms allow a given entity $y_i$ to be assigned to more than one cluster in $S$, as long as the assignments occur at different levels in the tree. This extra information regarding the relationships between clusters comes at a considerable cost, leading to a time complexity of $\mathcal{O}(N^2)$, or even $\mathcal{O}(N^3)$ depending on the actual algorithm in use \parencite{murtagh1984complexities,murtagh2011methods}. Partitional algorithms tend to converge in less time by comparison (details in Section \ref{Sec:KMeans}), but provide only information about the assignment of entities to clusters. Partitional algorithms were originally designed to produce a set of disjoint clusters, in which an entity $y_i \in Y$ could be assigned to a single cluster $S_k \in S$. K-Means \parencite{macqueen1967some,ball1967clustering,steinhaus1956division} is arguably the most popular of such algorithms (for more details see Section \ref{Sec:KMeans}). Among the many extensions to K-Means, we have Fuzzy C-Means \parencite{bezdek1981pattern} which applies Fuzzy set theory \parencite{zadeh1965fuzzy} to allow a given entity $y_i$ to be assigned to each cluster in $S$ at different degrees of membership. However, Fuzzy C-Means introduces other issues to clustering, falling outside the scope of this paper.

The popularity of K-Means is rather evident. A search in scholar.google.com for ``K-Means'' in May 2014 found just over $320,000$ results, the same search in May 2015 found $442,000$ results. Adding to these impressive numbers, implementations of this algorithm can be found in various software packages commonly used to analyse data, including SPSS, MATLAB, R, and Python. However, K-Means is not without weaknesses. For instance, K-Means treats every single feature in a data set equally, regardless of its actual degree of relevance. Clearly, different features in the same data set may have different degrees of relevance, a prospect we believe should be supported by any good clustering algorithm. With this weakness in mind research effort has happened over the last 30 years to develop K-Means based approaches supporting feature weighting (more details in Section \ref{Sec:MajorApproaches}). Such effort has lead to various different approaches, but unfortunately not much guidance on the choice of which to employ in practical applications.

In this paper, we provide the reader with a survey of K-Means based weighting algorithms. We find this survey to be unique because it does not simply explain some of the major approaches to feature weighting in K-Means, but also provides empirical evidence of their cluster recovery ability. We begin by formally introducing K-Means and the concept of feature weighting in Sections \ref{Sec:KMeans} and \ref{Sec:FeatureWeighting}, respectively. We then critically analyse some of the major methods for feature weighting in K-Means in Section \ref{Sec:MajorApproaches}. We chose to analyse those methods we believe are the most used or innovative, but since it is impossible to analyse all existing methods we are possibly guilty of omissions. The setting and results of our experiments can be found in Sections \ref{Sec:SettingOfExp} and \ref{Sec:Results}. The paper ends by presenting our conclusions and discussing common issues with these algorithms that could be addressed in future research, in Section \ref{Sec:Conclusion}.

\section{K-Means clustering}
\label{Sec:KMeans}

K-Means is arguably the most popular partitional clustering algorithm \parencite{jain2010data,steinley2006k,mirkin2012clustering}. For a given data set $Y$, K-Means outputs a disjoint set of clusters $S=\{S_1, S_2, ..., S_K\}$, as well as a centroid $c_k$ for each cluster $S_k \in S$. The centroid $c_k$ is set to have the smallest sum of distances to all $y_i \in S_k$, making $c_k$ a good general representation of $S_k$, often called a prototype. K-Means partitions a given data set $Y$ by iteratively minimising the sum of the within-cluster distance between entities $y_i \in Y$ and respective centroids $c_k \in C$. Minimising the equation below allows K-Means to show the natural structure of $Y$.
\begin{equation}
\label{Eq:KMeans}
W(S,C) = \sum_{k=1}^K \sum_{i \in S_k}\sum_{v \in V} (y_{iv} - c_{kv})^2,
\end{equation}
where $V$ represents the set of features used to describe each $y_i \in Y$. The algorithm used to iteratively minimise (\ref{Eq:KMeans}) may look rather simple at first, with a total of three steps, two of which iterated until the convergence. However, this minimisation is a non-trivial problem, being NP-Hard even if $K=2$ \parencite{aloise2009np}.
\begin{enumerate}
\itemsep0em
\item Select the values of $K$ entities from $Y$ as initial centroids $c_1, c_2, ..., c_K$. Set $S\leftarrow \emptyset$.
\item Assign each entity $y_i \in Y$ to the cluster $S_k$ represented by its closest centroid. If there are no changes in $S$, stop and output $S$ and $C$.
\item Update each centroid $c_k \in C$ to the centre of its cluster $S_k$. Go to Step 2.
\end{enumerate}

The K-Means criterion we show, (\ref{Eq:KMeans}), applies the squared Euclidean distance as in its original definition \parencite{macqueen1967some,ball1967clustering}. The use of this particular distance measure makes the centroid update in Step three of the algorithm above rather straightforward. Given a cluster $S_k$ with $|S_k|$ entities, $c_{kv} = \frac{1}{|S_k|}\sum_{i\in S_k}y_{iv}$, for each $v \in V$.

One can clearly see that K-Means has a strong relation with the Expectation Maximisation algorithm \parencite{dempster1977maximum}. Step two of K-Means can be seen as the expectation by keeping $C$ fixed and minimising (\ref{Eq:KMeans}) in respect to $S$, and Step three can be seen as the maximisation in which one fixes $S$ and minimises (\ref{Eq:KMeans}) in relation to $C$. K-Means also has a strong relation with Principal Component Analysis, the latter can be seen as a relaxation of the former \parencite{zha2001spectral,drineas2004clustering,ding2004k}.

K-Means, very much like any other algorithm in machine learning, has weaknesses. These are rather well-known and understood thanks to the popularity of this algorithm and the considerable research effort done by the research community. Among these weaknesses we have: (i) the fact that the number of clusters $K$ has to be known beforehand; (ii) K-Means will partition a data set $Y$ into $K$ partitions even if there is no clustering structure in the data; (iii) this is a greedy algorithm that may get trapped in local minima; (iv) the initial centroids, found at random in Step one heavily influence the final outcome; (v) it treats all features equally, regardless of their actual degree of relevance.

Here we are particularly interested in the last weakness. Regardless of the problem at hand and the structure of the data, K-Means treats each feature $v \in V$ equally. This means that features that are more relevant to a given problem may have the same contribution to the clustering as features that are less relevant. By consequence K-Means can be greatly affected by the presence of totally irrelevant features, including features that are solely composed of noise. Such features are not uncommon in real-world data. This weakness can be addressed by setting weights to each feature $v \in V$, representing its degree of relevance. We find this to be a particularly important field of research, we elaborate on the concept of feature weighting in the next section.

\section{Feature Weighting}
\label{Sec:FeatureWeighting}

New technology has made it much easier to acquire vast amounts of real-world data, usually described over many features. The \textit{curse of dimensionality} \parencite{bellman1956dynamic} is a term usually associated with the difficulties in analysing such high-dimensional data. As the number of features $v \in V$ increases, the minimum and maximum distances become impossible to distinguish as their difference, compared to the minimum distance, converges to zero \parencite{beyer1999nearest}. 
\begin{equation}
\lim_{|V| \to \infty} \frac{dist_{max} - dist_{min}}{dist_{min}} = 0
\end{equation} 
Apart from the problem above, there is a considerable consensus in the research community that meaningful clusters, particularly those in high-dimensional data, occur in subspaces defined by a specific subset of features \parencite{tsai2008developing,liu2005toward,chen2012feature,cordeiro2012minkowski}. In cluster analysis, and in fact any other pattern recognition task, one should not simply use all features available as clustering results become less accurate if a significant number of features are not relevant to some clusters \parencite{chan2004optimization}. Unfortunately, selecting the optimal feature subset is NP-Hard \parencite{blum1992training}.

Feature weighting can be thought of as a generalization of feature selection \parencite{wettschereck1997review,modha2003feature,tsai2008developing}. The latter has a much longer history and it is used to either select or deselect a given feature $v \in V$, a process equivalent to assigning a feature weight $w_v$ of one or zero, respectively. Feature selection methods effectively assume that each of the selected features has the same degree of relevance. Feature weighting algorithms do not make such assumption as there is no reason to believe that each of the selected features would have the same degree of relevance in all cases. Instead, such algorithms allow for a feature weight, normally in the interval $[0,1]$. This may be a feature weight $w_v$, subject to $\sum_{v \in V} w_v=1$, or even a cluster dependant weight $w_{kv}$, subject to $\sum_{v \in V} w_{kv}=1$ for $k = 1, 2, ..., K$. The idea of cluster dependant weights is well aligned with the intuition that a given feature $v$ may have different degrees of relevance at different clusters.

%More differences between feature weighting & selection HERE
Feature selection methods for unlabelled data follow either a filter or wrapper approach \parencite{dy2008unsupervised,kohavi1997wrappers}. The former uses properties of the data itself to select a subset of features during the data pre-processing phase. The features are selected before the clustering algorithm is run, making this approach usually faster. However, this speed comes at price. It can be rather difficult to define whether a given feature is relevant without applying clustering to the data. Methods following a wrapper approach make use of the information given by a clustering algorithm when selecting features. Often, these methods lead to better performance when compared to those following a filter approach \parencite{dy2008unsupervised}. However, these also tend to be more computationally intensive as the clustering and the feature selection algorithms are run. The surveys by \citeauthor{dy2008unsupervised} (\citeyear{dy2008unsupervised}), \citeauthor{steinley2008selection} (\citeyear{steinley2008selection}), and \citeauthor{guyon2003introduction} (\citeyear{guyon2003introduction}) are, in our opinion, a very good starting point for those readers in need of more information.
%%%%%%%%%%%%

Feature weighting and feature selection algorithms are not competing methods. The former does not dismiss the advantages given by the latter. Feature weighting algorithms can still deselect a given feature $v$ by setting its weight $w_v=0$, bringing benefits traditionally related to feature selection. Such benefits include those discussed by \citeauthor{guyon2003introduction} (\citeyear{guyon2003introduction}) and \citeauthor{dy2008unsupervised} (\citeyear{dy2008unsupervised}), such as a possible reduction in the feature space, reduction in measurement and storage requirements, facilitation of data understanding and visualization, reduction in algorithm utilization time, and a general improvement in cluster recovery thanks to the possible avoidance of the \textit{curse of dimensionality}.

Clustering algorithms recognise patterns under an unsupervised learning framework, it is only fitting that the selection or weighting of features should not require labelled samples. There are a considerable amount of unsupervised feature selection methods, some of which can be easily used in the data pre-processing stage \parencite{devaney1997efficient,talavera1999feature,mitra2002unsupervised} to either select or deselect features from $V$. Feature weighting algorithms for K-Means have thirty years of history, in the next section we discuss some what we believe to be the main methods. 

\section{Major approaches to feature weighting in K-Means}
\label{Sec:MajorApproaches}

Work on feature weighting in clustering has over 40 years of history \parencite{sneath1973numerical}, however, only in 1984 \parencite{desarbo1984synthesized} feature weighting was applied to K-Means, arguably the most popular partitional clustering algorithm. Many feature weighting algorithms based on K-Means have been developed since, here we chose nine algorithms for our discussion. These are either among the most popular, or introduce innovative new concepts.

%We describe all these algorithms using the same notation. We have a data set $Y$ composed of $N$ entities $y_i$, each described over the same features $v \in V$. These algorithms partition $Y$ into $K$ clusters $S=\{S_1, S_2, ..., S_K\}$, each represented by a centroid $c_k$, also described over the features $v \in V$. A given categorical feature $v$ has $L$ categories $a_v^1, a_v^2, ..., a_v^L$. The weight of feature $v$ is given by $w_v$, however, if this weight is cluster dependant, then $w_{kv}$ for a cluster $S_k$. Algorithms that group features $G = \{G_1, G_2, ..., G_t, ..., G_T\}$, if these have a weight, this is represented by $\omega_t$.

\subsection{SYNCLUS}
\label{Sec:SYNCLUS}
Synthesized Clustering (SYNCLUS) \parencite{desarbo1984synthesized} is, to our knowledge, the first K-Means extension to allow feature weights. SYNCLUS employs two types of weights by assuming that features, as well as groups of features, may have different degrees of relevance. This algorithm requires the user to meaningfully group features into $T$ partitions $G=\{G_1, G_2, ..., G_T\}$. We represent the degree of relevance of the feature group $G_t$ with $\omega_t$, where $1\leq t\leq T$. The feature weight of any given feature $v \in V$ is represented by $w_{v}$.

In its first step, very much like K-Means, SYNCLUS requests the user to provide a data set $Y$ and the desired number of partitions $K$. Unlike K-Means, the user is also requested to provide information regarding how the features are grouped, and a vector $\omega$  containing the weights of each feature group. This vector $\omega$ is normalised so that $\sum_t^T \omega_t=1$. DeSarbo suggests that each $w_{v}$, the weight of a given feature $v \in V$, can be initialised so that it is inversely proportional to the variance of $v$ over all entities $y_i \in Y$, or are all equal.

The distance between two objects $y_i$ and $y_j$ is defined, in each feature group, as their weighted squared distance $d(y_i, y_j)^{(t)}= \sum_{v \in G_t} w_{t_v}(y_{iv}-y_{jv})^2$. Given $\omega$, $w$, $Y$, $K$, and $d(y_i,y_j)^{(t)}$, for $i,j=1, 2, ..., N$, SYNCLUS optimises the weighted mean-square, stress-like objective function below. 

\begin{equation}
\label{Eq:SYNCLUS}
W(S,C,w,\omega)=\frac{\sum_t^T \omega_t \sum_{i \in Y} \sum_{j \in Y} (\delta_{ij}-d(y_i,y_j)^{(t)})}{\sum_{i \in Y} \sum_{j \in Y} \delta_{ij}^2},
\end{equation}
subject to a disjoint clustering so that $S_k \cap S_l=\emptyset$ for $k,l=1, 2, ..., K$ and $k\neq l$, as well as $\sum_{i \in Y} \sum_{j \in Y} \delta_{ij}^2 \neq 0$, $\delta_{ij}=\alpha a^*_{ij}+\beta$ (details regarding $\alpha$ and $\beta$ in \cite{desarbo1984synthesized}) where,
\begin{equation}
a^*_{ij}=
\begin{cases} 
\frac{1}{|S_k|},$ if $\{y_i, y_j\} \subseteq S_k,\\ 
0, $ otherwise$.
\end{cases}
\end{equation}
Although an icon of original research, SYNCLUS does have some weaknesses. This computationally expensive algorithm presented mixed results on empirical data sets \parencite{green1990preliminary}, and there have been other claims of poor performance \parencite{gnanadesikan1995weighting}. SYNCLUS is not appropriate for clusterwise regression context with both dependent and independent variables \parencite{desarbo1988maximum}. 

Nevertheless, SYNCLUS has been a target to various extensions. \citeauthor{desarbo1984constrained} (\citeyear{desarbo1984constrained}) extended this method to deal with constraints, different types of clustering schemes, as well as a general linear transformation of the features. It has also been extended by \citeauthor{makarenkov2001optimal} (\citeyear{makarenkov2001optimal}) by using the Polak-Ribiere optimisation procedure \parencite{polak1971computational} to minimise (\ref{Eq:SYNCLUS}). However, this latter extension seemed to be particularly useful only when `noisy' features (those without cluster structure) existed. The authors recommended using equal weights (ie. the original K-Means) when data are error-perturbed or contained outliers. 

The initial work on SYNCLUS also expanded into a method to find optimal feature weights for ultrametric and additive tree fitting \parencite{de1986optimal,de1988ovwtre}. However, this work lies outside the scope of this paper as the method was applied in hierarchical clustering.

SYNCLUS marked the beginning of research on feature weighting in K-Means, and it is possible to see its influences in nearly all other algorithms in this particular field.

\subsection{Convex K-Means}
\label{Sec:CK_Means}

\citeauthor{modha2003feature} (\citeyear{modha2003feature}) introduced the convex K-Means (CK-Means) algorithm. CK-Means presents an interesting approach to feature weighting by integrating multiple, heterogeneous feature spaces into K-Means. Given the two entities $\{y_i, y_j\} \subseteq Y$, each described over the features $v \in V$, the dissimilarity between these two entities is given by the distortion measure below.
\begin{equation}
\label{Eq:CK_Means_Distance}
D_w(y_i,y_j)=\sum_{v \in V} w_v D_v(y_{iv}, y_{jv}),
\end{equation}
where $D_v$ depends on the feature space in use. Modha and Spangler present two generic examples.
\begin{equation}
D_v(y_{iv}, y_{jv})=
\begin{cases} 
(y_{iv}-y_{jv})^T (y_{iv}-y_{jv}),$ in the Euclidean case$\\
2(1-y_{iv}^T y_{jv}), $ in the Spherical case$.
\end{cases}
\end{equation}
Equation (\ref{Eq:CK_Means_Distance}) allows calculating the distortion of a specific cluster $\sum_{y_i \in S_k}D_w(y_i, c_k)$, and the quality of the clustering $S=\{S_1, S_2, ..., S_K\}$, given by  $\sum_{k=1}^K\sum_{y_i \in S_k}D_w(y_i, c_k)$. CK-Means determines the optimal set of feature weights that simultaneously minimises the average within-cluster dispersion and maximises the average between-cluster dispersion along all of the feature spaces, by consequence minimising the criterion below.
\begin{equation}
\label{Eq:CK_Means_Criterion}
W(S,C,w)=\sum_{k=1}^K\sum_{y_i \in S_k} D_w(y_i, c_k).
\end{equation}
This method finds the optimal weight $w_v$ for each $v \in V$ from a pre-defined set of feature weights $\Delta =\{w:\sum_{v\in V} w_v=1, w_v\geq 0, v \in V\}$. Each partition $S^{(w)}=\{S_1^{(w)}, S_2^{(w)}, ..., S_K^{(w)}\}$ generated by minimising (\ref{Eq:CK_Means_Criterion}) with a different set of weights $w \in \Delta$ is then evaluated with a generalization of Fisher's discriminant analysis. In this, one aims to minimise the ratio between the average within-cluster distortion and the average between-cluster distortion. 

CK-Means can be thought of as a gradient descent method that never increases (\ref{Eq:CK_Means_Criterion}), and eventually converges to a local minima solution. This method has introduced a very interesting way to cluster entities described over different feature spaces, something we would dare say is a common characteristic of modern real-world data sets. CK-Means has also shown promising results in experiments \parencite{modha2003feature}, however, the way it finds feature weights has led to claims that generating $\Delta$ would be difficult in high-dimensional data \parencite{tsai2008developing,huang2005automated}, and that there is no guarantee the optimal weights would be in $\Delta$ \parencite{huang2005automated}.

\subsection{Attribute weighting clustering algorithm}
\label{Sec:Chan}

Another extension to K-Means to support feature weights was introduced by \citeauthor{chan2004optimization} (\citeyear{chan2004optimization}). This algorithm generates a weight $w_{kv}$ for each feature $v \in V$ at each cluster in $S=\{S_1, S_2, ..., S_k, ..., S_K\}$, within the framework of K-Means. This method supports the intuitive idea that different features may have different degrees of relevance at different clusters. This Attribute Weighting algorithm (AWK, for short) attempts to minimise the weighted squared distance between entities $y_i \in Y$ and their respective centroids $c_k \in C$, as per the criterion below.
\begin{equation}
\label{Eq:ChanWKMeans}
W(S,C,w)=\sum_{k=1}^K\sum_{i\in S_k}\sum_{v\in V} w_{kv}^{\beta} d(y_{iv}, c_{kv}),
\end{equation}
where $\beta$ is a user-defined parameter that is greater than 1, $d(y_{iv},c_{kv}) = |y_{iv} - c_{kv}|^2$ for a numerical $v$, and its the simple matching dissimilarity measure below for a categorical $v$.
\begin{equation}
\label{Eq:BasicCategorigalDistance}
d(y_{iv},c_{kv})=
\begin{cases} 
0$, if $y_{iv}=c_{kv}\\
1$, if $y_{iv}\neq c_{kv}.
\end{cases}
\end{equation}
The criterion (\ref{Eq:ChanWKMeans}) has a computational complexity complexity of $\mathcal{O}(NMK)$ \parencite{chan2004optimization}, where $M=|V|$ and is subject to:
\begin{enumerate}
\item A disjoint clustering, in which $S_k \cap S_l=\emptyset$ for $k,l=1, 2, ..., K$ and $k\neq l$.
\item A crisp clustering, given by $\sum_{k=1}^K |S_k| = N$.
\item $\sum_{v \in V} w_{kv}=1$ for a given cluster $S_k$.
\item $\{w_{kv}\}\geq 0$ for $k=1, 2, ..., K$ and $v \in V$.
\end{enumerate}
\citeauthor{chan2004optimization} (\citeyear{chan2004optimization}) minimises (\ref{Eq:ChanWKMeans}) under the above constraints by using partial optimisation for $S$, $C$ and $w$. The algorithm begins by setting each $w_{kv}=1/|V|$, fixing $C$ and $w$ in order to find the necessary conditions so $S$ minimises (\ref{Eq:ChanWKMeans}). Then one fixes $S$ and $w$, minimising (\ref{Eq:ChanWKMeans}) in respect to $C$. Next, one fixes $S$ and $C$ and minimises (\ref{Eq:ChanWKMeans}) in respect to $w$. This process is repeated until convergence. 

The minimisations of the first and second steps are rather straight forward. The assignment of entities to the closest cluster $S_k$ uses the weighted distance $d(y_i, c_k)=\sum_{v \in V} w_{kv} (y_{iv}-c_{kv})^2$, and since (\ref{Eq:ChanWKMeans}) clearly uses the squared Euclidean distance, $c_{kv}=\frac{1}{|S_k|} \sum_{i \in S_k} y_{iv}$. The minimisation of (\ref{Eq:ChanWKMeans}) is respect to $w$ depends on $\sum_{i \in S_k}(y_{iv} - c_{kv})^2$, generating the three possibilities below.
\begin{equation}
\label{Eq:ChanWKMeansWeights}
w_{kv}=
\begin{cases} 
\frac{1}{v^*},$ if $\sum_{i \in S_k}(y_{iv}-c_{kv})^2=0$, and $v^*=|\{v^{\prime}:\sum_{i \in S_k} (y_{iv^{\prime}}-c_{kv^{\prime}})^2=0\}|,\\
0,$ if $\sum_{i \in S_k}(y_{iv}-c_{kv})^2\neq 0$, but $\sum_{i \in S_k}(y_{iv^{\prime}}-c_{kv^{\prime}})^2=0, $ for some $v^{\prime} \in V,\\
\frac{1}{\sum_{j \in V}\left[\frac{\sum_{i \in S_k}(y_{iv} - c_{kv})^2}{\sum_{i \in S_k}(y_{ij} - c_{kj})^2}\right]^{\frac{1}{\beta -1}}}$,   if $\sum_{i \in S_k}(y_{iv}-c_{kv})^2\neq0.
\end{cases}
\end{equation}
The experiments in \citeauthor{chan2004optimization} (\citeyear{chan2004optimization}) deal solely with $\beta>1$. This is probably to avoid the issues related to divisions by zero that $\beta=1$ would present in (\ref{Eq:ChanWKMeansWeights}), and the behaviour of (\ref{Eq:ChanWKMeans}) at other values (for details see Section \ref{Sec:WK_Means}). It is interesting to see that \citeauthor{desarbo1984synthesized} (\citeyear{desarbo1984synthesized}) suggested two possible cases for initial weights in SYNCLUS (details in Section \ref{Sec:SYNCLUS}), either to set all weights to the same number, or to be inversely proportional to the variance of the feature in question. It seems to us that Chan's method have used both suggestion, by initializing each weight $w_{kv}$ to $1/|V|$ and by optimising $w_{kv}$ so that it is higher when the dispersion of $v$ in $y_{iv} \in S_k$ is lower, as the third case in (\ref{Eq:ChanWKMeansWeights}) shows.

There are some issues to have in mind when using this algorithm. The use of (\ref{Eq:BasicCategorigalDistance}) may be problematic in certain cases as the range of $d(y_{iv},c_{kv})$ will be different depending on whether $v$ is numerical or categorical. Based on the work of Huang (\citeyear{huang1998extensions}) and \citeauthor{ng2002clustering} (\citeyear{ng2002clustering}), \citeauthor{chan2004optimization} introduces a new parameter to balance the numerical and categorical parts of a mixed data set, in an attempt to avoid favouring either part. In their paper they test AWK using different values for this parameter and the best is determined as that resulting in the highest cluster recovery accuracy. This approach is rather hard to follow in real-life clustering scenarios as no labelled data would be present. This approach was only discussed in the experiments part of the paper, not being present in the AWK description so it is ignored in our experiments.

Another point to note is that their experiments using real-life data sets, despite all explanations about feature weights, use two weights for each feature. One of these relates to the numerical features while the other relates to those that are categorical. This approach was also not explained in the AWK original description and is ignored in our experiments as well.

A final key issue to this algorithm, and in fact various others, is that there is no clear method to estimate the parameter $\beta$. Instead, the authors state that their method is not sensitive to a range of values of $\beta$, but unfortunately this is demonstrated with experiments on synthetic data in solely two real-world data sets.

\subsection{Weighted K-Means}
\label{Sec:WK_Means}
\citeauthor{huang2005automated} (\citeyear{huang2005automated}) introduced the Weighted K-Means (WK-Means) algorithm. WK-Means attempts to minimise the object function below, which is similar to that of \citeauthor{chan2004optimization} (\citeyear{chan2004optimization}), discussed in Section \ref{Sec:Chan}. However, unlike the latter, WK-Means originally sets a single weight $w_v$ for each feature $v \in V$.
\begin{equation}
\label{Eq:WKMeans}
W(S,C,w)=\sum_{k=1}^K\sum_{i\in S_k}\sum_{v\in V} w_{v}^{\beta} d(y_{iv}, c_{kv}),
\end{equation}
The Equation above is minimised using an iterative method, optimising (\ref{Eq:WKMeans}) for $S$, $C$, and $w$, one at a time. During this process Huang et al. presents the two possibilities below for the update of $w_v$, with $S$ and $C$ fixed, subject to $\beta>1$.
\begin{equation}
w_v=
\begin{cases} 
0$, if $D_v=0\\
\frac{1}{\sum_{j=1}^h \frac{D_v}{D_j}^{\frac{1}{\beta-1}}}$, if $D_v \neq 0,
\end{cases}
\end{equation}
where,
\begin{equation}
D_v=\sum_{k=1}^K \sum_{i \in S_k} d(y_{iv}, c_{kv}),
\end{equation}
and $h$ is the number of features where $D_v \neq 0$. If $\beta=1$, the minimisation of (\ref{Eq:WKMeans}) follows that $w_{v^\prime}=1$, and $w_v=0$, where $v^\prime \neq v$, and $D_{v^\prime} \leq D_v$, for each $v \in V$ \parencite{huang2005automated}.

The weight $w_v^{\beta}$ in (\ref{Eq:WKMeans}) makes the final clustering $S$, and by consequence the centroids in $C$, dependant of the value of $\beta$. There are two possible critical values for $\beta$, $0$ and $1$. If $\beta = 0$, Equation (\ref{Eq:WKMeans}) becomes equivalent to that of K-Means (\ref{Eq:KMeans}). At $\beta=1$, the weight of a single feature $v \in V$ is set to one (that with the lowest $D_v$), while all the others are set to zero. Setting $\beta =1$ is probably not desirable in most problems.

The above critical values generate three intervals of interest. When $\beta <0$, $w_v$ increases with an increase in $D_v$. However, the negative exponent makes $w_v^{\beta}$ smaller, so that $v$ has less of an impact on distance calculations. If $0<\beta<1$, $w_v$ increases with an increase in $D_v$, so does $w_v^{\beta}$. This goes against the principle that a feature with a small dispersion should have a higher weight, proposed by \citeauthor{chan2004optimization} (\citeyear{chan2004optimization}) (perhaps inspired by SYNCLUS, see Section \ref{Sec:SYNCLUS}), and followed by \citeauthor{huang2005automated} (\citeyear{huang2005automated}). If $\beta >1$, $w_v$ decreases with an increase in $D_v$, and so does $w_v^{\beta}$, very much the desired effect of decreasing the impact of a feature $v$ in (\ref{Eq:WKMeans}) whose $D_v$ is high.

WK-Means was later extended to support fuzzy clustering \parencite{li2006novel}, as well as cluster dependant weights \parencite{huang2008weighting}. The latter allows WK-Means to support weights with different degrees of relevance at different clusters, each represented by $w_{kv}$. This required a change in the criterion to be minimised to $W(S,C,w)=\sum_{k=1}^K\sum_{i\in S_k}\sum_{v\in V} w_{kv}^{\beta} d(y_{iv}, c_{kv})$, and similar changes to other related equations.

In this new version, the dispersion of a variable $v \in V$ at a cluster $S_k$ is given by $D_{kv}=\sum_{i \in S_k} (d(y_{iv}, c_{kv}) + c)$, where $c$ is a  user-defined constant. The authors suggest that in practice $c$ can be chosen as the average dispersion of all features in the data set. More importantly, the adding of $c$ addresses a considerable shortcoming. A feature whose dispersion $D_{kv}$ in a particular cluster $S_k$ is zero should not be assigned a weight of zero when in fact $D_{kv}=0$ indicates that $v$ may be an important feature to identify cluster $S_k$. An obvious exception is if $\sum_{k=1}^K D_{kv} = 0$ for a given $v$, however, such feature should normally be removed in the data pre-processing stage.

Although there have been improvements, the final clustering is still highly dependant on the exponent exponent $\beta$. It seems to us that the selection of $\beta$ depends on the problem at hand, but unfortunately there is no clear strategy for its selection. We also find that the lack of relationship between $\beta$ and the distance exponent (two in the case of the Euclidean squared distance) avoids the possibility of seen the final weights as feature re-scaling factors. Finally, although WK-Means supports cluster-dependant features, all features are treated as if they were a homogeneous feature space, very much unlike CK-Means (details in Section \ref {Sec:CK_Means}).

\subsection{Entropy Weighting K-Means}
\label{Sec:EWKM}

The Entropy Weighting K-Means algorithm (EW-KM) \parencite{jing2007entropy} minimises the within cluster dispersion while maximising the negative entropy. The reasoning behind this is to stimulate more dimensions to contribute to the identification of clusters in high-dimensional sparse data, avoiding problems related to identifying such clusters using only a few dimensions.

With the above in mind, \citeauthor{jing2007entropy} (\citeyear{jing2007entropy}) devised the following criterion for EW-KM:
\begin{equation}
\label{EQ:EW_KM}
W(S,C,w) = \sum_{k=1}^K \left[\sum_{i \in S_k}^N \sum_{v \in V} w_{kv}(y_{iv} - c_{kv})^2 + \gamma \sum_{v \in V} w_{kv} log w_{kv} \right],
\end{equation}
subject to $\sum_{v \in V} w_{kv} = 1$, $\{w_{kv}\} \geq 0$, and a crisp clustering. In the criterion above, one can easily identify that the first term is the weighted sum of the within cluster dispersion. The second term, in which $\gamma$ is a parameter controlling the incentive for clustering in more dimensions, is the negative weight entropy.

The calculation of weights in EW-KM occurs as an extra step in relation to K-Means, but still with a time complexity of $\mathcal{O}(rNMK)$ where $r$ is the number of iterations the algorithm takes to converge. Given a cluster $S_k$, the weight of each feature $v \in V$ is calculated one at a time with the equation below.
\begin{equation}
w_{kv} = \frac{exp(\frac{-D_{kv}}{\gamma})}{\sum_{j \in V} exp(\frac{-D_{kj}}{\gamma})},
\end{equation}
where $D_{kv}$ represents the dispersion of feature $v$ in the cluster $S_k$, given by $D_{kv} = \sum_{i \in S_k} (y_{iv} - c_{kv})^2$. As one would expect, the minimisation of (\ref{EQ:EW_KM}) uses partial optimisation for $w$, $C$, and $S$. First, $C$ and $w$ are fixed and (\ref{EQ:EW_KM}) is minimised in respect to $S$. Next, $S$ and $w$ are fixed and (\ref{EQ:EW_KM}) is minimised in respect to $C$. In the final step, $S$ and $C$ are fixed, and (\ref{EQ:EW_KM}) is minimised in respect to $w$. This adds a single step to K-Means, used to calculate feature weights.

The R package \textit{weightedKmeans} found at CRAN includes an implementation of this algorithm, which we decided to use in our experiments (details in Sections \ref{Sec:SettingOfExp} and \ref{Sec:Results}). \citeauthor{jing2007entropy} (\citeyear{jing2007entropy}) presents extensive experiments, with synthetic and real-world data. These experiments show EW-KM outperforming various other clustering algorithms. However, there are a few points we should note. First, it is somewhat unclear how a user should choose a precise value for $\gamma$. Also, most of the algorithms used in the comparison required a parameter as well. Although we understand it would be too laborious to analyse a large range of parameters for each of these algorithms, there is no much indication on reasoning behind the choices made.

\subsection{Improved K-Prototypes}
\label{Sec:IKP}

\citeauthor{ji2013improved} (\citeyear{ji2013improved}) have introduced the Improved K-Prototypes clustering algorithm (IK-P), which minimises the WK-Means criterion (\ref{Eq:WKMeans}), with influences from k-prototype \parencite{huang1998extensions}. IK-P introduces the concept of distributed centroid to clustering, allowing the handling of categorical features by adjusting the distance calculation to take into account the frequency of each category.

IK-P treats numerical and categorical features differently, but it is still able to represent the cluster $S_k$ of a data set $Y$ containing mixed type, data with a single centroid $c_k=\{c_{k1}, c_{k2}, ..., c_{k|V|}\}$. Given a numerical feature $v$, $c_{kv}= \frac{1}{|S_k|} \sum_{i \in S_k} y_{iv}$, the center given by the Euclidean distance. A categorical feature $v$ containing $L$ categories $a \in v$, has $c_{kv}=\{\{a_v^1, \omega_{kv}^1\}, \{a_v^2, \omega_{kv}^2\}, ..., \{a_v^l, \omega_{kv}^l\}, ..., \{a_v^L, \omega_{kv}^L\}\}$. This representation for a categorical $v$ allows each category $a \in v$ to have a weight $\omega_{kv}^l=\sum_{i \in S_k}\eta(y_{iv})$, directly related to its frequency in the data set $Y$.
\begin{equation}
\eta(y_{iv})=
\begin{cases} 
\frac{1}{\sum_{i\in S_k} 1}$, if $y_{iv}=a_v^l,\\
0$, if $y_{iv}\neq a_v^l.
\end{cases}
\end{equation}
Such modification also requires a re-visit of the distance function in (\ref{Eq:WKMeans}). The distance is re-defined to the below.
\begin{equation}
\label{Eq:IKP_Distance}
d(y_{iv}, c_{kv})=
\begin{cases} 
|y_{iv} - c_{kv}|$, if $v$ is numerical,$\\
\varphi(y_{iv} - c_{kv})$, if $v$ is categorical$,
\end{cases}
\end{equation}
where $\varphi(y_{iv} - c_{kv}) = \sum_{k=1}^K \vartheta(y_{iv}, a_v^l)$,
\begin{equation}
\vartheta(y_{iv}, a_v^k)=
\begin{cases} 
0$, if $y_{iv}=a_v^l,\\
\omega_{iv}^k$, if $y_{iv} \neq a_v^l,
\end{cases}
\end{equation}

IK-P presents some very interesting results \parencite{ji2013improved}, outperforming other popular clustering algorithms such as k-prototype, SBAC, and KL-FCM-GM \parencite{chatzis2011fuzzy,ji2012fuzzy}. However, the algorithm still leaves some open questions. 

For instance, \citeauthor{ji2013improved} (\citeyear{ji2013improved}) present experiments on six data sets (two of which being different versions of the same data set) setting $\beta=8$, but it is not clear whether the good results provided by this particular $\beta$ would generalize to other data sets. Given a numerical feature, IK-P applies the Manhattan distance (\ref{Eq:IKP_Distance}), however, centroids are calculated using the mean. The center of the Manhattan distance is given by the median rather than the mean, this is probably the reason why Ji et al. found it necessary to allow the user to set a maximum numbers of iterations to their algorithm. Now, even if the algorithm converges, most likely it would converge in a smaller number of iterations if the distance used for the assignments of entities was aligned to that used for obtaining the centroids. Finally, while $d(y_{iv},c_{kv})$ for a categorical $v$ has a range in the interval $[0,1]$, the same is not true if $v$ is numerical, however, \citeauthor{ji2013improved} (\citeyear{ji2013improved}) make no mention to data standardization.
\subsection{Intelligent Minkowski Weighted K-Means}

Previously, we have extended WK-Means (details in Section \ref{Sec:WK_Means}) by introducing the intelligent Minkowski Weighted K-Means (iMWK-Means) \parencite{cordeiro2012minkowski}. In its design, we aimed to propose a deterministic algorithm supporting non-elliptical clusters with weights that could be seen as feature weighting factors. To do so, we combined the Minkowski distance and intelligent K-Means \parencite{mirkin2012clustering}, a method that identifies anomalous patterns in order find the number of clusters in a data set, as well as good initial centroids.

Below, we show the Minkowski distance between the entities $y_i$ and $y_j$, described over features $v \in V$.
\begin{equation}
\label{Eq:MinkowskiDistance}
d(y_i,y_j)=(\sum_{v \in V} |y_{iv} - y_{jv}|^p)^{1/p},
\end{equation}
where $p$ is a user-defined parameter. If $p$ equals $1, 2,$ or $ \infty$, Equation (\ref{Eq:MinkowskiDistance}) is equivalent to the the Manhattan, Euclidean and  Chebyshev distances, respectively. Assuming a given data set has two dimensions (for easy visualisation), the distance bias of a clustering algorithm using (\ref{Eq:MinkowskiDistance}) would be towards clusters whose shape are any interpolation between a diamond ($p=1$) and a square ($p=\infty$), clearly going through a circle ($p=2$). This is considerably more flexible than algorithms based solely on the squared Euclidean distance, as these recover clusters biased towards circles only. One can also see the Minkowski distance as a multiple of the power mean of the feature-wise differences between $y_i$ and $y_j$.

The iMWK-Means algorithm calculates distances using (\ref{Eq:WeightedMinkDist}), a weighted version of the $p^{th}$ root of (\ref{Eq:MinkowskiDistance}). The use of a root is analogous to the frequent use of the squared Euclidean distance in K-Means.
\begin{equation}
\label{Eq:WeightedMinkDist}
d(y_i,y_j)=\sum_{v \in V} w_{kv}^p |y_{iv} - y_{jv}|^p,
\end{equation}
where the user-defined parameter $p$ scales the distance as well as well as the cluster dependent weight $w_{kv}$. This way the feature weights can be seen as feature re-scaling factors, this is not possible for WK-Means when $\beta \neq 2$. Re-scaling a data set with these feature re-scaling factors increases the likelihood of various cluster validity indices to lead to the correct number of clusters \parencite{de2015recovering}. With (\ref{Eq:WeightedMinkDist}) one can reach the iMWK-Means criterion below.
\begin{equation}
W(S,C,w)=\sum_{k=1}^K\sum_{i\in S_k}\sum_{v\in V} w_{kv}^p |y_{iv} - c_{kv}|^p.
\end{equation}
The update of $w_{kv}$, for each $v \in V$ and $k=1, 2, ..., K$, follows the equation below.
\begin{equation}
w_{kv}=\frac{1}{\sum_{u \in V} \frac{D_{kvp}}{D_{kup}}^\frac{1}{p-1}},
\end{equation}
where the dispersion of feature $v$ in cluster $k$ is now dependant on the exponent $p$, $D_{kvp}=\sum_{i \in S_k} |y_{iv}-c_{kv}|^p + c$, and $c$ is a constant equivalent to the average dispersion. The update of the centroid of cluster $S_k$ on feature $v$, $c_{kv}$ also depends on the value of $p$. At values of $p$ $1, 2,$ and $\infty$, the center of (\ref{Eq:MinkowskiDistance}) is given by the median, mean and midrange, respectively. If $p \notin \{1,2,\infty\}$ then the center can be found using a steepest descend algorithm \parencite{cordeiro2012minkowski}.

The iMWK-Means algorithm deals with categorical features by transforming them in numerical, following a method described by \citeauthor{mirkin2012clustering} (\citeyear{mirkin2012clustering}). In this method, a given categorical feature $v$ with $L$ categories is replaced by $L$ binary features, each representing one of the original categories. For a given entity $y_i$, only the binary feature representing $y_{iv}$ is set to one, all others are set to zero. The concept of distributed centroid \parencite{ji2013improved} can also be applied to our algorithm \parencite{amorimandmakarenkov2015applying}, however, in order to show a single version of our method we decided not to follow the latter here.

Clearly, the chosen value of $p$ has a considerable impact on the final clustering given by iMWK-Means. \citeauthor{cordeiro2012minkowski} (\citeyear{cordeiro2012minkowski}) introduced a semi-supervised algorithm to estimate a good $p$, requiring labels for 20\% of the entities in $Y$. Later, the authors showed that it is indeed possible to estimate a good value for $p$ using only 5\% of labelled data under the same semi-supervised method, and presented a new unsupervised method to estimate $p$, requiring no labelled samples \parencite{amorim2014selectingmink}.

The iMWK-Means proved to be superior to various other algorithms, including WK-Means with cluster dependant weights \parencite{cordeiro2012minkowski}. However, iMWK-Means also has room for improvement. Calculating a centroid for a $p \notin \{1, 2, \infty\}$ requires the use of a steepest descent method. This can be time consuming, particularly when compared with other algorithms defining $c_{kv} = \frac{1}{|S_k|} \sum_{i \in S_k} y_{iv}$. Although iMWK-Means allows for a distance bias towards non-elliptical clusters, by setting $p \neq 2$, it assumes that all clusters should be biased towards the same shape. 
\subsection{Feature Weight Self-Adjustment K-Means}

\citeauthor{tsai2008developing} (\citeyear{tsai2008developing}) integrated a feature weight self-adjustment mechanism (FWSA) to K-Means. In this mechanism finding $w_v$ for $v \in V$ is modelled as an optimisation problem to simultaneously minimise the separations within clusters and maximise the separation between clusters. The former is measured $a_v=\sum_{k=1}^K \sum_{i \in S_k} d(y_{iv}, c_{kv})$, where $d()$ is a function returning the distance between the feature $v$ of entity $y_i$ and that of centroid $c_k$. The separation between clusters of a given feature $v$ is measured by $b_v=\sum_{k=1}^K N_k d(c_{kv}, c_v)$, where $N_k$ is the number of entities in $S_k$, and $c_v$ is the center of feature $v$ over $y_i \in Y$. With $a_v$ and $b_v$ one can evaluate how much the feature $v$ contributes to the clustering quality, and in a given iteration $j$ calculate $w_v^{(j+1)}$.
\begin{equation}
w_v^{(j+1)}=\frac{1}{2} (w_v^{(j)} + \frac{b_v^{(j)}/a_v^{(j)}}{\sum_{v\in V}(b_v^{(j)}/a_v^{(j)})}),
\end{equation}
where the multiplication by $1/2$ makes sure that $w_v \in [0,1]$ so it can satisfy the constrain $\sum_{v\in V}w_v^{(j+1)}=1$. With $w_v$ one can then minimise the criterion below.
\begin{equation}
W(S,C,w)=\sum_{k=1}^K\sum_{i\in S_k}\sum_{v\in V} w_{v} (y_{iv}- c_{kv})^2,
\end{equation}
subject to $\sum_{v \in V} w_v=1$ and $\{w_v\}_{v \in V} \geq 0$, and a crisp clustering. Experiments in synthetic and real-world data sets compare FWSA K-Means favourably in relation to WK-Means. However, it assumes a homogeneous feature space, and like the previous algorithms it still evaluates a single feature at a time. This means that that a group of features, each irrelevant on its own, but informative if in a group, would be discarded.

FWSA has already been compared to WK-Means in three real-world data sets \parencite{tsai2008developing}. This comparison shows FWSA reaching an adjusted Rand index (ARI) 0.77 when applied to the Iris data set, while WK-Means reached only 0.75. The latter ARI was obtained by setting $\beta=6$, but our experiments (details in Section \ref{Sec:Results}) show that WK-Means may reach 0.81. The difference may be related to how the data was standardised, as well as how $\beta$ was set as here we found the best at each run. Of course, the fact that FWSA does not require a user-defined parameter is a considerable advantage.

\subsection{FG-K-Means}
\label{Sec:FGK}

%%%
%Many high-dimensional data sets are the result of integration of measurements on observations from different perspectives so that features of different measurements can be grouped.
%%%

The FG-K-Means (FGK) algorithm \parencite{chen2012feature} extends K-Means by applying weights at two levels, features and clusters of features. This algorithm has been designed to deal with large data sets whose data comes from multiple sources. Each of the $T$ data sources provides a subset of features $G=\{G_1, G_2, ..., G_T\}$, where $G_t \neq \emptyset$, $G_t \subset V$, $G_t \cap G_s = \emptyset$ for $t \neq s$ and $1 \leq t, s \leq T$, and $\cup G_t = V$. Given a cluster $S_k$, FGK identifies the degree of relevance of a feature $v$, represented by $w_{kv}$, as well as the relevance of a group of features $G_t$, represented by $\omega_{kt}$. Unlike SYNCLUS (details in Section \ref{Sec:SYNCLUS}), FGK does not require the weights of the groups of features to be entered by the user. FGK updates the K-Means criterion (\ref{Eq:KMeans}) to include both $w_{kv}$ and $\omega_{kt}$, as we show below.
\begin{equation}
\label{Eq:FGK}
W(S,C,w,\omega)= \sum_{k=1}^K \left[ \sum_{i \in S_k} \sum_{t=1}^T \sum_{v \in G_t} \omega_{kt} w_{kv} d(y_{iv}, c_{kv}) + \lambda \sum_{t=1}^T \omega_{kt} log(\omega_{kt}) + \eta \sum_{v \in V} w_{kv} log(w_{kv})\right],
\end{equation}
where $\lambda$ and $\eta$ are user-defined parameters, adjusting the distributions of the weights related to the groups of features in $G$, and each of the features $v \in V$, respectively. The minimisation of (\ref{Eq:FGK}) is subject to a crisp clustering in which any given entity $y_i \in Y$ is assigned to a single cluster $S_k$. The feature group weights are subject to $\sum_{k=1}^K \omega_{kt}=1$, $0 < \omega_{kt} < 1$, for $1 \leq t \leq T$. The feature weights are subject to $\sum_{v \in G_t} w_{kv}=1$, $0 < w_{kv} < 1$, for $1 \leq k \leq K$ and $1 \leq t \leq T$.

Given a numerical $v$, the function $d$ in (\ref{Eq:FGK}) returns the squared Euclidean distance between $y_{iv}$ and $c_{kv}$, given by $(y_{iv} - c_{kv})^2$. A categorical $v$ leads to $d$ returning one if $y_{iv} = c_{kv}$, and zero otherwise, very much like (\ref{Eq:BasicCategorigalDistance}). The update of each feature weight follows the equation below.

\begin{equation}
w_{kv}=\frac{exp(\frac{-E_{kv}}{\eta})}{\sum_{h \in G_t} exp( \frac{-E_{kh}}{\eta})},
\end{equation}
where $ E_{kv}=\sum_{i \in S_k} \omega_{kt} d(y_{iv}, c_{kv})$, and $t$ is the index of the feature group to which feature $v$ is assigned to. The update of the feature group weights follows.

\begin{equation}
\omega_{kt} = \frac{exp(\frac{-D_{kt}}{\lambda})}{\sum_{s=1}^T exp(\frac{-D_{ks}}{\lambda})},
\end{equation}
where, $D_{kt} = \sum_{i \in S_k} \sum_{v \in G_t} w_{kv} d(y_{iv}, c_{kv})$. Clusterings generated by FGK are heavily dependant on $\lambda$ and $\eta$. These parameters must set to positive real values. Large values for $\lambda$ and $\eta$ lead to weights to be more evenly distributed, so more subspaces contribute to the clustering. Low values lead to weights being more concentrated on fewer subspaces, each of these having a larger contribution to the clustering.

FGK has a time complexity of $\mathcal{O}(rNMK)$, where $r$ is the number of iterations this algorithm takes to complete \parencite{chen2012feature}. It has outperformed K-Means, WK-Means (see Section \ref{Sec:WK_Means}), LAC  \parencite{domeniconi2007locally} and EW-KM (see Section \ref{Sec:EWKM}), but it also introduces new open questions. This particular method was designed aiming to deal with high-dimensional data, however, it is not clear how $\lambda$ and $\eta$ should be estimated. This issue makes it rather hard to use FGK in real-world problems. We find that it would also be interesting to see a generalization of this method to use other distance measures, allowing a different distance bias.

\section{Setting of the experiments}
\label{Sec:SettingOfExp}

In our experiments we have used real-world as well as synthetic data sets. The former were obtained from the popular UCI machine learning repository \parencite{Lichman:2013} and include data sets with
different combinations of numerical and categorical features, as we show in Table \ref{Tab:ReadlDataNoNoise}.
\begin{table}[h]\small
\caption{Real-world data sets used in our comparison experiments.}
\begin{center}
\tabcolsep=0.11cm
\begin{tabular}{lccccc}
&Entities&Clusters&\multicolumn{3}{c}{Original features}\\
\cline{4-6}
&&&Numerical&Categorical&Total\\
Australian credit&690&2&6&8&14\\
Balance scale&625&2&0&4&4\\
Breast cancer &699&2&9&0&9\\
Car evaluation&1728&4&0&6&6\\
Ecoli&336&8&7&0&7\\
Glass&214&6&9&0&9\\
Heart (Statlog)&270&2&6&7&13\\
Ionosphere&351&2&33&0&33\\
Iris&150&3&4&0&4\\
Soybean&47&4&0&35&35\\
Teaching Assistant&151&3&1&4&5\\
Tic Tac Toe&958&2&0&9&9\\
Wine&178&3&13&0&13\\
\end{tabular}
\end{center}
\label{Tab:ReadlDataNoNoise}
\end{table}

The synthetic data sets contain spherical Gaussian clusters so that the covariance matrices are diagonal with the same diagonal value $\sigma^2$ generated at each cluster randomly between $0.5$ and $1.5$. All centroid components were generated independently from a Gaussian distribution with zero mean and unity variance. The cardinality of each cluster was generated following an uniformly random distribution, constrained to a minimum of 20 entities. We have generated 20 data sets under each of the following configurations: (i) 500x4-2, 500 entities over four features partitioned into two clusters; (ii) 500x10-3, 500 entities over 10 features partitioned into three clusters; (iii) 500x20-4, 500 entities over 20 features partitioned into four clusters; (iv) 500x50-5, 500 entities over 50 features partitioned into five clusters. 

Unfortunately, we do not know the degree of relevance of each feature in all of our data sets. For this reason we decided that our experiments should also include data sets to which we have added noise features. Given a data set, for each of its categorical features we have added a new feature composed entirely of uniform random integers (each integer simulates a category). For each of its numerical features we have added a new feature composed entirely of uniform random values. In both cases the new noise feature has the same domain as the original feature. This approach has effectively doubled the number of features in each data set, as well as the number of data sets used in our experiments.

Prior to our experiments we have standardised the numerical features of each of our data sets as per the equation below.
%`
\begin{equation}
\label{Eq:Stand}
y_{iv} = \frac{y_{iv}- \overline{y_v}}{range(y_v)},
\end{equation}
where $\overline{y_v} = \frac{1}{N}\sum_{i=1}^N y_{iv}$, and $range(y_v)=max(\{y_{iv}\}_{i=1}^N)-min(\{y_{iv}\}_{i=1}^N)$. Our choice of using (\ref{Eq:Stand}) instead of the popular \textit{z}-score is perhaps easier to explain with an example. Lets imagine two features, unimodal $v_1$ and multimodal $v_2$. The standard deviation of $v_2$ would be higher than that of $v_1$ which means that the \textit{z}-score of $v_2$ would be lower. Thus, the contribution of $v_2$ to the clustering would be lower than that of $v_1$ even so it is $v_2$ that has a cluster structure. Arguably, a disadvantage of using (\ref{Eq:Stand}) is that it can be detrimental to algorithms based on other standardisation methods \parencite{steinley2008new}, not included in this paper.

We have standardised the categorical features for all but those experiments with the Attribute weighting and Improved K-Prototypes algorithms (described in Sections \ref{Sec:Chan} and \ref{Sec:IKP}, respectively). These two algorithms define distances for categorical features, so transforming the latter to numerical is not required. Given a categorical feature $v$ containing $q$ categories we substitute $v$ by $q$ new binary features. In a given entity $y_i$ only a single of these new features is set to one, that representing the category originally in $y_{iv}$. We then numerically standardise each of the new features by subtracting it by its mean. The mean of a binary feature is in fact its frequency, so a binary feature representing a category with a high frequency contributes less to the clustering than one with a low frequency.
In terms of data pre-processing, we also made sure that all features in each data set had a range higher than zero. Features with a range of zero are not meaningful so they were removed.

Unfortunately, we found it very difficult to set a fair comparison including all algorithms we describe in this paper. SYNCLUS and FG-K-Means(described in Sections \ref{Sec:SYNCLUS} and \ref{Sec:FGK}, respectively) go a step further in feature weighting by allowing weights for feature groups. However, they both require the user to meaningfully group features $v \in V$ into $T$ partitions $G=\{G_1, G_2, ..., G_T\}$ with SYNCLUS also requesting the user to set a weight for each of these groups. Even if we had enough information to generate $G$ it would be unfair to provide this extra information to some algorithms and not to others. If we were to group features randomly we would be providing these two algorithms with misleading information more often than not, which would surely have an impact on their cluster recovery ability. If we were to set a single group of features and give this group a weight of one then we would be removing the main advantage of using these algorithms, and in fact FG-K-Means would be equivalent to EW-KM. Convex K-Means (Section \ref{Sec:CK_Means}) also goes a step further in feature weighting, it does so by integrates multiple, heterogeneous feature spaces. \citeauthor{modha2003feature} (\citeyear{modha2003feature}) demonstrates that with the Euclidean and Spherical cases. However, there is little information regarding the automatic detection of the appropriate space given a feature $v \in V$, a very difficult problem indeed. For these reasons we decided not to include these three algorithms in our experiments.

\section{Results and discussion}
\label{Sec:Results}

In our experiments we do have a set of labels for each data set. This allows us to measure the cluster recovery of each algorithm in terms of the adjusted Rand index (ARI) \parencite{hubert1985comparing} between the generated clustering and the known labels.
\begin{equation}
ARI = \frac{ \sum_{ij} \binom{n_{ij}}{2} - [\sum_i \binom{a_i}{2} \sum_j \binom{b_j}{2}] / \binom{n}{2} }{ \frac{1}{2} [\sum_i \binom{a_i}{2} + \sum_j \binom{b_j}{2}] - [\sum_i \binom{a_i}{2} \sum_j \binom{b_j}{2}] / \binom{n}{2} },
\end{equation}
where $n_{ij}=|S_i \cap S_j|$, $a_i = \sum_{j=1}^K |S_i \cap S_j|$ and $b_i = \sum_{i=1}^K |S_i \cap S_j|$.

FWSA is the only algorithm we experiment with that does not require an extra parameter from the user. This is clearly an important advantage as estimating optimum parameters is not a trivial problem, and in many cases the authors of the algorithms do not present a clear estimation method. 

The experiments we show here do not deal with parameter estimation. Instead, we determine the optimum parameter for a given algorithm by experimenting with values from $1.0$ to $5.0$ in steps of $0.1$. The only exception are the experiments with EW-KM where we apply values from $0$ to $5.0$ in steps of $0.1$, this is because EW-KM is the only algorithm in which a parameter between zero and one is also appropriate. Most of the algorithms we experiment with are non-deterministic (iMWK-Means is the only exception), so we run each algorithm at each parameter $100$ times and select as the optimum parameter that with the highest average ARI.

Tables \ref{Tab:ExpRealDataPart1NoNoise} and \ref{Tab:ExpRealDataPart2NoNoise} show the results of our experiments on the real-world data sets without noise features added to them (see Table \ref{Tab:ReadlDataNoNoise}). We show the average (together with the standard deviation) and maximum ARI for what we found to be the optimum parameter for each data set we experiment with. There are different comparisons we can make, particularly when one of the algorithms is deterministic, iMWK-Means. If we compare the algorithms in terms of their expected ARI, given a good parameter, then we can see that in 9 data sets iMWK-Means reaches the highest ARI. We run each non-deterministic algorithm 100 times for each parameter value. If we take into account solely the highest ARI over these 100 runs then the EW-KM reaches the highest ARI overall in 9 data sets, while IK-P does so in six. Still looking only at the highest ARI, the FWSA algorithm (the only algorithm not to require an extra parameter) reaches the highest ARI in four data sets, the same number as AWK and WK-Means. Another point of interest is that the best parameter we could find for iMWK-Means was the same in four data sets.
\begin{table}[h]
\caption{Experiments with the real-world data sets with no added noise features. The standard deviation can be found after the backslash under the mean. Par refers to the parameter value required by the algorithm.}
\begin{center}
\tabcolsep=0.11cm
\begin{tabular}{lcccccccccccc}
&\multicolumn{3}{c}{Attribute Weighting}&&\multicolumn{3}{c}{Weighted K-Means}&&\multicolumn{3}{c}{Entropy WK}\\
\cline{2-4}
\cline{6-8}
\cline{10-12}
&\multicolumn{2}{c}{ARI}&&&\multicolumn{2}{c}{ARI}&&&\multicolumn{2}{c}{ARI}\\
\cline{2-3}
\cline{6-7}
\cline{10-11}
&Mean&Max&Par&&Mean&Max&Par&&Mean&Max&Par\\
Australian&0.15/0.11&0.50&4.9&&0.19/0.23&0.50&2.2&&0.31/0.19&0.50&0.90\\ 
Balance&0.03/0.03&0.19&3.4&&0.04/0.04&0.18&4.7&&0.04/0.05&0.23&2.10\\ 
Breast c.&0.68/0.20&0.82&4.5&&0.83/0.00&0.83&2.6&&0.85/0.01&0.87&1.10\\ 
Car eva.&0.07/0.05&0.22&2.6&&0.04/0.05&0.13&1.2&&0.07/0.06&0.22&4.60\\ 
Ecoli&0.02/0.02&0.04&2.5&&0.42/0.06&0.57&3.5&&0.45/0.09&0.72&0.10\\ 
Glass&0.15/0.03&0.22&3.7&&0.19/0.04&0.28&2.8&&0.17/0.03&0.28&0.10\\ 
Heart&0.21/0.16&0.45&4.7&&0.18/0.07&0.27&1.1&&0.33/0.10&0.39&3.10\\ 
Ionosphere&0.14/0.08&0.25&1.2&&0.18/0.05&0.34&1.2&&0.18/0.00&0.21&0.70\\ 
Iris&0.80/0.16&0.89&1.6&&0.81/0.11&0.89&3.9&&0.71/0.14&0.82&0.30\\ 
Soybean&0.57/0.18&1.00&3.5&&0.78/0.22&1.00&3.1&&0.74/0.20&1.00&0.10\\ 
Teaching A.&0.02/0.01&0.05&1.9&&0.02/0.01&0.07&4.0&&0.03/0.02&0.10&0.20\\ 
Tic Tac Toe&0.02/0.03&0.07&1.4&&0.03/0.04&0.15&4.1&&0.02/0.03&0.15&1.00\\ 
Wine&0.76/0.06&0.82&4.8&&0.85/0.02&0.90&4.4&&0.82/0.05&0.90&0.30\\ 
\end{tabular}
\end{center}
\label{Tab:ExpRealDataPart1NoNoise}
\end{table}
\begin{table}[h]
\caption{Experiments with the real-world data sets with no added noise features. The standard deviation can be found after the backslash under the mean. Par refers to the parameter value required by the algorithm. IMWK-Means is a deterministic algorithm and FWSA does not require a parameter, hence the dashes.}
\begin{center}
\tabcolsep=0.11cm
\begin{tabular}{lccccccccccc}
&\multicolumn{3}{c}{Improved K-P}&&\multicolumn{3}{c}{Intelligent Minkowski WK}&&\multicolumn{3}{c}{Feature Weight Self Adj.}\\
\cline{2-4}
\cline{6-8}
\cline{10-12}
&\multicolumn{2}{c}{ARI}&&&\multicolumn{2}{c}{ARI}&&&\multicolumn{2}{c}{ARI}\\
\cline{2-3}
\cline{6-7}
\cline{10-11}
&Mean&Max&Par&&Mean&Max&Par&&Mean&Max&Par\\
Australian&0.15/0.08&0.20&4.7&&-&0.50&1.1&&0.20/0.21&0.50&-\\ 
Balance&0.04/0.05&0.23&1.3&&-&0.09&3.3&&0.03/0.03&0.15&-\\ 
Breast c.&0.74/0.00&0.74&4.9&&-&0.85&4.6&&0.81/0.12&0.83&-\\ 
Car eva.&0.03/0.05&0.22&5.0&&-&0.13&2.0&&0.04/0.06&0.22&-\\
Ecoli&0.46/0.00&0.46&3.0&&-&0.04&2.5&&0.37/0.06&0.52&-\\ 
Glass&0.21/0.06&0.31&4.4&&-&0.28&4.6&&0.16/0.04&0.25&-\\
Heart&0.31/0.08&0.36&4.6&&-&0.31&2.9&&0.15/0.10&0.31&-\\ 
Ionosphere&0.14/0.07&0.43&1.9&&-&0.21&1.1&&0.17/0.03&0.21&-\\ 
Iris&0.78/0.21&0.90&1.2&&-&0.90&1.1&&0.77/0.19&0.89&-\\ 
Soybean&0.87/0.16&0.95&2.4&&-&1.00&1.8&&0.71/0.23&1.00&-\\ 
Teaching A.&0.01/0.01&0.04&4.0&&-&0.04&2.2&&0.02/0.01&0.05&-\\ 
Tic Tac Toe&0.02/0.03&0.15&2.8&&-&0.02&1.1&&0.02/0.02&0.15&-\\ 
Wine&0.86/0.01&0.86&4.3&&-&0.82&1.6&&0.70/0.13&0.82&-\\
\end{tabular}
\end{center}
\label{Tab:ExpRealDataPart2NoNoise}
\end{table}

Tables \ref{Tab:ExpRealDataPart1WithNoise} and \ref{Tab:ExpRealDataPart2WithNoise} show the results of our experiments on the real-world data sets with noise features added to them. Given a data set $Y$, for each $v \in V$ we add a new feature to $Y$ composed entirely of uniform random values (integers in the case of a categorical $v$) with the same domain as $v$. This effectively doubles the cardinality of $V$. In this set of experiments the Improved K-Prototype was unable to find eight clusters in the Ecoli data set. We believe this issue is related to the data spread. The third feature of this particular data set has only 10 entities with a value other than 0.48. The fourth feature has a single entity with a value other than 0.5. Clearly on the top of these two issues we have an extra seven noise features. Surely one could argue that features three and four could be removed from the data set as they are unlikely to be informative. However, we decided not to start opening concessions to algorithms. Instead we expect the algorithms to find these issues and deal with them. This turn, when comparing expected ARI values given a good parameter, iMWK-Means reaches the highest ARI value in 8 data sets. It ceased to reach the highest ARI in the Australian data set in which it now reaches 0.22 while EW-KM reaches 0.34 (that is 0.3 more than in the experiments with no noise features, but such small inconsistencies are to be expected in experiments with non-deterministic algorithms). When comparing the maximum possible for each algorithm the WK-Means algorithm does reach the highest ARI in six data sets, while EW-KM does so in five.

\begin{table}[h]
\caption{Experiments with the real-world data sets with added noise features. The standard deviation can be found after the backslash under the mean. Par refers to the parameter value required by the algorithm.}
\begin{center}
\tabcolsep=0.11cm
\begin{tabular}{lccccccccccc}
&\multicolumn{3}{c}{Attribute Weighting}&&\multicolumn{3}{c}{Weighted K-Means}&&\multicolumn{3}{c}{Entropy WK}\\
\cline{2-4}
\cline{6-8}
\cline{10-12}
&\multicolumn{2}{c}{ARI}&&&\multicolumn{2}{c}{ARI}&&&\multicolumn{2}{c}{ARI}\\
\cline{2-3}
\cline{6-7}
\cline{10-11}
&Mean&Max&Par&&Mean&Max&Par&&Mean&Max&Par\\
Australian&0.16/0.07&0.22&4.8&&0.21/0.23&0.50&1.3&&0.34/0.17&0.50&5.00\\ 
Balance&0.01/0.02&0.11&2.4&&0.02/0.03&0.18&3.2&&0.02/0.03&0.13&0.30\\ 
Breast c.&0.32/0.00&0.32&1.8&&0.84/0.00&0.84&4.9&&0.86/0.00&0.87&1.70\\ 
Car eva.&0.05/0.05&0.14&2.6&&0.03/0.04&0.14&3.0&&0.04/0.05&0.15&2.80\\ 
Ecoli&0.01/0.01&0.04&3.8&&0.38/0.08&0.50&1.2&&0.34/0.05&0.42&0.20\\ 
Glass&0.16/0.03&0.24&5.0&&0.20/0.04&0.27&1.1&&0.14/0.04&0.22&0.30\\ 
Heart&0.20/0.14&0.41&4.8&&0.22/0.12&0.33&1.2&&0.36/0.09&0.45&0.20\\ 
Ionosphere&0.19/0.07&0.27&1.2&&0.18/0.03&0.21&1.2&&0.17/0.02&0.18&0.20\\ 
Iris&0.77/0.17&0.89&4.4&&0.79/0.12&0.87&1.5&&0.64/0.08&0.73&0.20\\ 
Soya&0.46/0.16&0.94&4.0&&0.76/0.21&1.00&1.5&&0.61/0.21&1.00&0.30\\ 
Teaching A.&0.02/0.01&0.07&4.8&&0.01/0.01&0.08&2.0&&0.02/0.01&0.05&0.30\\ 
Tic Tac Toe&0.03/0.03&0.07&1.6&&0.02/0.03&0.15&2.3&&0.02/0.02&0.10&0.80\\ 
Wine&0.76/0.07&0.88&3.8&&0.84/0.02&0.87&1.5&&0.77/0.03&0.82&0.10\\ 
\end{tabular}
\end{center}
\label{Tab:ExpRealDataPart1WithNoise}
\end{table}
\begin{table}[h]
\caption{Experiments with the real-world data sets with added noise features. The standard deviation can be found after the backslash under the mean. Par refers to the parameter value required by the algorithm. IMWK-Means is a deterministic algorithm and FWSA does not require a parameter, hence the dashes.}
\begin{center}
\tabcolsep=0.11cm
\begin{tabular}{lccccccccccc}
&\multicolumn{3}{c}{Improved K-P}&&\multicolumn{3}{c}{Intelligent Minkowski WK}&&\multicolumn{3}{c}{Feature Weight Self Adj.}\\
\cline{2-4}
\cline{6-8}
\cline{10-12}
&\multicolumn{2}{c}{ARI}&&&\multicolumn{2}{c}{ARI}&&&\multicolumn{2}{c}{ARI}\\
\cline{2-3}
\cline{6-7}
\cline{10-11}
&Mean&Max&Par&&Mean&Max&Par&&Mean&Max&Par\\
Australian&0.15/0.09&0.20&4.9&&-&0.22&1.7&&0.18/0.22&0.50&-\\ 
Balance&0.02/0.04&0.13&1.4&&-&0.08&2.8&&0.01/0.02&0.09&-\\
Breast c.&0.73/0.00&0.73&4.5&&-&0.87&1.4&&0.65/0.34&0.83&-\\ 
Car eva.&0.02/0.03&0.12&1.1&&-&0.04&2.5&&0.03/0.05&0.14&-\\ 
Ecoli&-&-&-&&-&0.04&2.5&&0.09/0.09&0.29&-\\ 
Glass&0.23/0.05&0.30&3.4&&-&0.23&2.5&&0.10/0.06&0.21&-\\ 
Heart&0.32/0.06&0.36&4.0&&-&0.30&3.9&&0.10/0.10&0.35&-\\ 
Ionosphere&0.12/0.04&0.38&2.1&&-&0.29&1.5&&0.16/0.05&0.21&-\\ 
Iris&0.82/0.12&0.85&3.2&&-&0.90&1.1&&0.75/0.28&0.89&-\\ 
Soya&0.90/0.11&0.95&2.1&&-&1.00&1.4&&0.67/0.19&1.00&-\\ 
Teaching A.&0.00/0.01&0.02&4.7&&-&0.05&2.9&&0.01/0.01&0.05&-\\ 
Tic Tac Toe&0.02/0.04&0.15&1.1&&-&0.07&1.1&&0.02/0.03&0.15&-\\
Wine&0.83/0.03&0.90&4.4&&-&0.83&1.2&&0.55/0.25&0.81&-\\ 
\end{tabular}
\end{center}
\label{Tab:ExpRealDataPart2WithNoise}
\end{table}

Tables \ref{Tab:ExpGMMPart1} and \ref{Tab:ExpGMMPart2} show the results of our experiments on the synthetic data sets with and without noise features. Given a data set $Y$, for each $v \in V$ we have added a new feature to $Y$ containing uniformly random noise in the same domain as that of $v$, very much like what we did in the real-world data sets. The only difference is that in the synthetic data sets we do not have categorical features and we know that they contain Gaussian clusters (see Section \ref{Sec:SettingOfExp}). We have 20 data sets for each of the data set configurations, hence, the values under max represent the average of the maximum ARI obtained in each of the 20 data sets, as well as the standard deviation of these values.  

In this set of experiments iMWK-Means reached the highest expected ARI in all data sets, with and without noise features. If we compare solely the maximum possible ARI per algorithm WK-Means reaches the highest ARI in three data set configurations with no noise features added to them, and in two of the data sets with noise features. AWK also reaches the highest ARI in two of the configurations
\begin{table}[h]\scriptsize
\caption{Experiments with the synthetic data sets, with and without noise features. The standard deviation can be found after the backslash under the mean. Par refers to the parameter value required by the algorithm.}
\begin{center}
\tabcolsep=0.09cm
\begin{tabular}{lccccccccccc}
&\multicolumn{3}{c}{Attribute Weighting}&&\multicolumn{3}{c}{Weighted K-Means}&&\multicolumn{3}{c}{Entropy WK}\\
\cline{2-4}
\cline{6-8}
\cline{10-12}
&\multicolumn{2}{c}{ARI}&&&\multicolumn{2}{c}{ARI}&&&\multicolumn{2}{c}{ARI}\\
\cline{2-3}
\cline{6-7}
\cline{10-11}
&Mean&Max&Par&&Mean&Max&Par&&Mean&Max&Par\\
No noise\\
500x4-2&0.50/0.36&0.61/0.31&4.11/1.21&&0.61/0.32&0.62/0.31&3.28/1.26&&0.62/0.30&0.66/0.26&2.35/1.15\\%&&0.45/0.36&0.69/0.22&3.42/0.71\\ 
500x10-3&0.62/0.20&0.83/0.11&4.55/0.53&&0.68/0.20&0.85/0.10&3.10/1.05&&0.67/0.20&0.84/0.10&0.83/0.39\\%&&0.40/0.24&0.80/0.15&1.29/0.15\\ 
500x20-4&0.74/0.22&0.98/0.02&4.09/0.66&&0.75/0.25&0.99/0.02&3.11/1.02&&0.75/0.24&0.98/0.03&0.35/0.22\\%&&0.44/0.22&0.90/0.08&0.90/0.42\\
500x50-5&0.83/0.18&1.00/0.01&3.48/1.03&&0.82/0.19&1.00/0.00&3.62/0.99&&0.80/0.19&1.00/0.00&0.24/0.12\\ %&&0.51/0.20&0.92/0.06&1.70/1.57\\
\cline{1-12}
With noise\\
500x4-2&0.29/0.38&0.60/0.32&2.93/1.11&&0.27/0.37&0.61/0.32&1.46/0.80&&0.34/0.33&0.56/0.29&0.29/0.16\\ %&&0.38/0.37&0.68/0.24&3.16/0.39\\ 
500x10-3&0.59/0.20&0.80/0.13&4.19/0.78&&0.61/0.23&0.85/0.10&1.29/0.15&&0.54/0.25&0.78/0.14&0.32/0.26\\ %&&0.27/0.22&0.66/0.17&1.42/0.24\\
500x20-4&0.73/0.22&0.98/0.03&3.78/0.90&&0.71/0.25&0.93/0.19&1.37/0.26&&0.81/0.14&0.95/0.06&0.34/0.10\\%&&0.28/0.18&0.69/0.12&0.90/0.09\\
500x50-5&0.83/0.18&1.00/0.01&3.14/0.81&&0.82/0.20&0.98/0.10&2.01/1.22&&0.84/0.15&1.00/0.01&0.52/0.41\\%&&0.32/0.17&0.71/0.11&0.55/0.05\\
\end{tabular}
\end{center}
\label{Tab:ExpGMMPart1}
\end{table}
\begin{table}[h]\small
\caption{Experiments with the synthetic data sets, with and without noise features. The standard deviation can be found after the backslash under the mean. Par refers to the parameter value required by the algorithm. IMWK-Means is a deterministic algorithm and FWSA does not require a parameter, hence the dashes.}
\begin{center}
\tabcolsep=0.10cm
\begin{tabular}{lccccccccccc}
&\multicolumn{3}{c}{Improved K-P}&&\multicolumn{3}{c}{Intelligent Minkowski WK}&&\multicolumn{3}{c}{Feature Weight Self Adj.}\\
\cline{2-4}
\cline{6-8}
\cline{10-12}
&\multicolumn{2}{c}{ARI}&&&\multicolumn{2}{c}{ARI}&&&\multicolumn{2}{c}{ARI}\\
\cline{2-3}
\cline{6-7}
\cline{10-11}
&Mean&Max&Par&&Mean&Max&Par&&Mean&Max&Par\\
No noise\\
500x4-2&0.45/0.37&0.59/0.32&3.69/1.18&&-&0.63/0.30&3.18/1.31&&0.38/0.36&0.57/0.33&-\\
500x10-3&0.60/0.21&0.81/0.12&3.85/0.85&&-&0.71/0.18&2.51/0.83&&0.42/0.23&0.67/0.24&-\\
500x20-4&0.74/0.25&0.98/0.04&3.74/1.03&&-&0.90/0.17&2.58/1.05&&0.64/0.24&0.94/0.15&-\\ 
500x50-5&0.82/0.19&1.00/0.00&3.50/1.14&&-&1.00/0.01&1.73/0.94&&0.77/0.20&0.97/0.11&-\\ 
\cline{1-12}
With noise\\
500x4-2&0.27/0.36&0.58/0.33&2.47/1.05&&-&0.48/0.40&1.79/1.25&&0.02/0.13&0.47/0.37&-\\ 
500x10-3&0.55/0.22&0.79/0.13&2.84/1.01&&-&0.85/0.09&1.58/0.26&&0.07/0.19&0.39/0.35&-\\ 
500x20-4&0.71/0.25&0.93/0.15&2.58/0.92&&-&0.95/0.06&1.80/0.69&&0.26/0.30&0.88/0.22&-\\
500x50-5&0.82/0.20&1.00/0.01&2.65/0.96&&-&0.94/0.08&2.24/0.88&&0.72/0.26&0.97/0.12&-\\ 
\end{tabular}
\end{center}
\label{Tab:ExpGMMPart2}
\end{table}
Clearly, there are other comparisons we can make using all algorithms described in Section \ref{Sec:MajorApproaches}. Based on the information we present in Section \ref{Sec:MajorApproaches} about each algorithm, as well as the cluster recovery results we present in this section, we have defined eight characteristics we believe are desirable for any K-Means based clustering algorithm that implements feature weighting. Table \ref{Tab:Comparison8Characteristics} shows our comparison, which we now describe one characteristic at a time.

\textit{No extra user-defined parameter}. Quite a few of the algorithms we describe in Section \ref{Sec:MajorApproaches} require an extra parameter to be defined by the user. By tuning this parameter (or these parameters, in the case of FGK) each of these algorithms is able to achieve high accuracy in terms of cluster recovery. However, it seems to us that this parameter estimation is a non-trivial task, particularly because the optimum value is problem dependant. This makes it very difficult to suggest a generally good parameter value (of course this may not be the case if one knows how the data is distributed). Since different values for a parameter tend to result in different clusterings, one could attempt to estimate the best clustering by applying a clustering validation index \parencite{arbelaitz2013extensive,amorim2014selectingmink}, consensus clustering \parencite{goder2008consensus}, or even a semi-supervised learning approach \parencite{cordeiro2012minkowski}. Regarding the latter, we have previously demonstrated that with as low as 5\% of the data being labelled it is still possible to estimate a good parameter for iMWK-Means \parencite{amorim2014selectingmink}.

\textit{It is deterministic}. A K-Means generated clustering heavily depends on the initial centroids this algorithm uses. These initial centroids are often found at random, meaning that if K-Means is run twice, it may generate very different clusterings. It is often necessary to run this algorithm a number of times and then somehow identify which clustering is the best (again, perhaps using a clustering validation index, a consensus approach, or in the case of this particular algorithm the output of its criterion). If a K-Means based feature weighting algorithm is also non-deterministic, chances are one will have to determine the best parameter and then the best run when applying that parameter. One could also run the algorithm many times per parameter and apply a clustering validation index to each of the generated clusterings. In any case, this can be a very computationally intensive task. We find it that the best approach would be to have a feature weighting algorithm that is deterministic, requiring the algorithm to be run a single time. The iMWK-Means algorithm applies a weighted Minkowski metric based version of the intelligent K-Means \parencite{mirkin2012clustering}. The latter algorithm finds anomalous clusters in a given data set and uses this information to generate initial centroids, making iMWK-Means a deterministic algorithm.

\textit{Accepts different distance bias}. Any distance in use will lead to a bias in the clustering. For instance, the Euclidean distance is biased towards spherical shapes while the Manhattan distance is biased towards diamond shapes. A good clustering algorithm should allow for the alignment of its distance bias to the data at hand. Two of the algorithms we analyse address this issue, but in very different ways. CK-Means is able to integrate multiple, heterogeneous feature spaces into K-Means, this means that each feature may use a different distance measure, and by consequence have a different bias. This is indeed a very interesting, and intuitive approach, as features measure different things so they may be in different spaces. The iMWK-Means also allows for different distance bias, it does so by using the $L_p$ metric, leaving the exponent $p$ as a user-defined parameter (see Equation \ref{Eq:WeightedMinkDist}). Different values for the exponent $p$ lead to different distance biases. However, this algorithm still assumes that all clusters in the data set have the same bias.

\textit{Supports at least two weights per feature}. In order to model the degree of relevance of a particular feature one may need more than a single weight. There are two very different cases that one should take into consideration: (i) a given feature $v \in V$ may be considerably informative when attempting to discriminate a cluster $S_k$, but not so for other clusters. This leads to the intuitive idea that $v$ should in fact have $K$ weights. This approach is followed by AWK, WK-Means (in its updated version, see \cite{huang2008weighting}), EWK-Means, iMWK-Means and FGK; (ii) a given feature $v \in V$ may be not be, on its own, informative to any cluster $S_k \in S$. However, the same feature may be informative when grouped with other features. Generally speaking, two (or more) features that are useless by themselves may be useful together \parencite{guyon2003introduction}. FGK is the only algorithm we analyse that calculates weights for groups of features.

\textit{Features grouped automatically}. If a feature weighting algorithm should take into consideration the weights of groups of features, it should also be able to group features on its own. This is probably the most controversial of the characteristics we analyse because none of the algorithms we deal with here is able to do so. We present this characteristic in Table \ref{Tab:Comparison8Characteristics} to emphasise its importance. Both algorithms that deal with weights for groups of features, SYNCLUS and FGK, require the users to group the features themselves. We believe that perhaps an approach based on bi-clustering \parencite{mirkin1998mathematical} could address this issue.

\textit{Calculates all used feature weights}. This is a basic requirement of any feature weighting algorithm. It should be able to calculate all feature weights it needs. Of course a given algorithm may support initial weights being provided by the user, but it should also be able to optimise these if needed. SYNCLUS requires the user to input the weights for groups of features and does not optimise these. CK-Means requires all possible weights to be put in a set  $\Delta =\{w:\sum_{v\in V} w_v=1, w_v\geq 0, v \in V\}$ and then tests each possible subset of $\Delta$, the weights are not calculated. This approach can be very time consuming, particularly in high-dimensional data.

\textit{Supports categorical features}. Data sets often contain categorical features. These features may be transformed to numerical values, however, such transformation may lead to loss of information and considerable increase in dimensionality. Most of the analysed algorithms that support categorical features do so by setting a simple matching dissimilarity measure (eg. AWK, WK-Means and FGK). This binary dissimilarity is zero iff both features have exactly the same category (see for instance Equation \ref{Eq:BasicCategorigalDistance}), and one otherwise. IK-P presents a different and interesting approach taking into account the frequency of each category at a categorical $v$. This allows for a continuous dissimilarity measure in the interval [0,1].

\textit{Analyses groups of features}. Since two features that are useless by themselves may be useful together \parencite{guyon2003introduction}, a feature weighting algorithm should be able to calculate weights for groups of features. Only a single algorithm we have analysed is able to do so, FGK. SYNCLUS also uses weights for groups of features, however, these are input by the user rather than calculated by the algorithm.
\begin{table}[h]\small
\caption{A comparison of the discussed feature weighting algorithms over eight key characteristics.}
\begin{center}
\begin{tabular}{lcccccccc}
&\begin{sideways}No extra user-defined parameter\end{sideways}&\begin{sideways}It is deterministic\end{sideways}&\begin{sideways}Accepts different distance bias\end{sideways}&\begin{sideways}Supports at least two weights per feature\end{sideways}&\begin{sideways}Features grouped automatically\end{sideways}&\begin{sideways}Calculates all used feature weights\end{sideways}&\begin{sideways}Supports categorical features\end{sideways}&\begin{sideways}Analyses groups of features\end{sideways}\\
\cline{2-9}
SYNCLUS&\Checkmark&&&\Checkmark\\
CK-Means&\Checkmark&&\Checkmark\\
AWK&&&&\Checkmark&&\Checkmark&\Checkmark\\
\cline{1-9}
WK-Means&&&&\Checkmark &&\Checkmark&\Checkmark\\
EWK-Means&&&&\Checkmark&&\Checkmark\\
IK-P&&&&&&\Checkmark&\Checkmark\\
\cline{1-9}
iMWK-Means&&\Checkmark&\Checkmark&\Checkmark&&\Checkmark\\
FWSA&\Checkmark &&&&&\Checkmark \\
FGK&&&&\Checkmark&&\Checkmark&\Checkmark&\Checkmark\\
\end{tabular}
\end{center}
\label{Tab:Comparison8Characteristics}
\end{table}

\section{Conclusion and future directions}
\label{Sec:Conclusion}

Recent technology has made it incredibly easy to acquire vast amounts of real-world data. Such data tend to be described in high-dimensional spaces, forcing data scientists to address difficult issues related to the \textit{curse of dimensionality}. Dimensionality reduction in machine learning is commonly done using feature selection algorithms, in most cases during the data pre-processing stage. This type of algorithm can be very useful to select relevant features in a data set, however, they assume that all relevant features have the same degree of relevance, which is often not the case.

Feature weighting is a generalisation of feature selection. The former models the degree of relevance of a given feature by giving it a weight, normally in the interval $[0,1]$. Feature weighting algorithms can also deselect a feature, very much like feature selection algorithms, by simply setting its weight to zero. K-Means is arguably the most popular partitional clustering algorithm. Efforts to integrate feature weighting in K-Means have been done for the last 30 years (for details, see Section \ref{Sec:MajorApproaches}). 

In this paper we have provided the reader with a discussion on nine of the most popular or innovative feature weighting mechanisms for K-Means. Our survey also presents an empirical comparison including experiments in real-world and synthetic data sets, both with and without noise features. Because of the difficulties of presenting a fair empirical comparison (see Section \ref{Sec:SettingOfExp}) we experimented with six of the nine algorithms discussed. Our survey shows some issues that are somewhat common in these algorithms and could be addressed in future research. For instance, each of the algorithms we discuss presents at least one of the following issues: 

(i) the criterion to be minimised includes a new parameter (or more), but unfortunately there is no clear strategy for the selection of a precise value for this parameter. This issue applies to most algorithms we discussed. Future research could address this issue in different ways. For instance, a method could use one or more clustering validation indices (for a recent comparison of these, see \cite{arbelaitz2013extensive}) to measure the quality of clusterings obtained applying different parameter values. It could also apply a consensus clustering based approach \parencite{goder2008consensus}, assuming that two entities that should belong to the same cluster are indeed clustered together by a given algorithm more often than not, over different parameter values. methods developed in future research could also apply a semi-supervised approach, this could require as low as 5\% of the data being labelled in order to estimate a good parameter \parencite{amorim2014selectingmink}.

(ii) the method treats all features as if they were in the same feature space, often not the case in real-world data. CK-Means is an exception to this rule, it integrates multiple, heterogeneous feature spaces. It would be interesting to see this idea expanded in future research to other feature weighting algorithms. Another possible approach to this issue would be to measure dissimilarities using different distance measures but compare them using a comparable scale, for instance the distance scaled by the sum of the data scatter. Of course this could lead to new problems, such as for instance defining what distance measure should be used at each feature. 

(iii) the method assumes that all clusters in a given data set should have the same distance bias. It is intuitive that different clusters in a given data set may have different shapes. However, in the algorithms we discuss when a dissimilarity measure is chosen it introduces a shape bias that is the same for all clusters in the data set. Future research could address this issue by allowing different distance measures at different clusters, leading to different shape biases. However, this could be difficult to achieve given what we argue in (ii) and that one would need to align each cluster to the bias of a distance measure.

(iv) features are evaluated one at a time, presenting difficulties for cases when the discriminatory information is present in a group of features, but not in any single feature of this group. In order to deal with this issue a clustering method should be able to group such features and calculate a weight for the group. Perhaps the concept of bi-clustering \parencite{mirkin1998mathematical} could be extended in future research by clustering features and entities, but also weighting features and groups of features.

The above ideas for future research address indeed some of the major problems we have today in K-Means based feature weighting algorithms. Of course this does not mean they are easy to implement, in fact we acknowledge quite the opposite.
\printbibliography
\end{document}